%% file: ewerton_iros_2021.tex
\documentclass[letterpaper, 10 pt, conference]{ieeeconf}  

\UseRawInputEncoding

\IEEEoverridecommandlockouts                              

\usepackage{amsmath} 
\usepackage{amssymb}  
\usepackage{siunitx}

\usepackage{float}
\usepackage[dvips]{graphicx}
\usepackage[dvips]{graphics}
\usepackage{graphics} 
\usepackage{epsfig} 
\usepackage{times} 
\usepackage{ctable}
\usepackage{stfloats}
\usepackage[caption=false]{subfig}
\usepackage{multirow}
\usepackage[caption=false]{subfig}
\usepackage{booktabs}
\usepackage{tabu}
\usepackage[T1]{fontenc}
\usepackage{url}
\newcommand\jmonote[1]{}
\title{\LARGE \bf
An Efficient Image-to-Image Translation HourGlass-based Architecture \\ for Object Pushing Policy Learning 
}

\author{Marco Ewerton, Angel Mart\'inez-Gonz\'alez, Jean-Marc Odobez 
\thanks{The authors are with the Idiap Research Institute, CH-1920, Martigny, Switzerland
		{\tt\small \{marco.ewerton, angel.martinez, jean-marc.odobez\}@idiap.ch}
}
\thanks{The research leading to these results has received funding from the Swiss National Science Foundation through the HEAP project (Human-Guided Learning and Benchmarking of Robotic Heap Sorting, ERA-net CHIST-ERA).}}

\newcommand{\mypartitle}[2][2.3]{\vspace*{-#1 ex}~\\{\noindent {\bf #2}}}

\input{notation}

\begin{document}

\maketitle
\thispagestyle{empty}
\pagestyle{empty}

\begin{abstract}

  Humans effortlessly solve pushing tasks in everyday life but unlocking these capabilities remains a challenge in robotics because physics models of these tasks are often inaccurate or unattainable.
  State-of-the-art data-driven approaches learn to compensate for these inaccuracies or replace the approximated physics models altogether.
  Nevertheless, approaches like Deep Q-Networks (DQNs) suffer from local optima in large state-action spaces. Furthermore, they rely on well-chosen deep learning architectures and learning paradigms.
  In this paper, we propose to frame the learning of pushing policies (where to push and how) by DQNs as an image-to-image translation problem and exploit an Hourglass-based architecture.
  We present an architecture combining a predictor of which pushes lead to changes in the environment with a state-action value predictor dedicated to the pushing task. Moreover, we investigate positional information encoding to learn position-dependent policy behaviors.
  We demonstrate in simulation experiments with a UR5 robot arm that our overall architecture helps the DQN learn
  faster and achieve higher performance in a pushing task involving objects with unknown dynamics. Code available at \url{https://github.com/idiap/hourglass_push}
\end{abstract}

\input{introduction}


\input{architecture}



\input{experiments}


\section{Conclusion}
\label{sec:conclusion}
In this paper, we presented an approach for object pushing policy learning from 
visual data.
We have addressed the task as an image-to-image translation and
proposed to use Hourglass architectures to produce high-resolution state-action value maps given an input depth image.
We divided the task into valid pushing spot detection and pushing state-action value prediction
subtasks, and designed their rewards so that pushing actions effectively 
produce changes in the environment and solve the task at hand, in this case pushing objects into a box.
Our results suggest that the use of the Hourglass architecture is better suitable
for predicting more accurate pushing actions and is more efficient than other
classification-based architectures such as DenseNet.


\addtolength{\textheight}{-12cm}   





\bibliographystyle{IEEEtran}
\bibliography{ewerton_iros_2021}

\end{document}

%% file: notation.tex
\usepackage{xspace}

\newcommand{\Baseline}{{\sc DenseNet}\xspace}
\newcommand{\HourGlass}{{\sc HGlassN}\xspace}
\newcommand{\HourGlassLPE}{{\sc HGlassN-Pos}\xspace}
\newcommand{\WithMask}{{\sc Mask}\xspace}
\newcommand{\WithRL}{{\sc OnlineRL}\xspace}

\newcommand{\NObjBox}{\ensuremath{N_{ObjB}}\xspace}
\newcommand{\NObjGround}{\ensuremath{N_{ObjG}}\xspace}
\newcommand{\NObjTable}{\ensuremath{N_{ObjT}}\xspace}
\newcommand{\Naction}{\ensuremath{N_{Act}}\xspace}

\newcommand{\MaskNet}{{\em PushMask}\xspace}
\newcommand{\RewardNet}{{\em PushReward}\xspace}
\newcommand{\PushNet}{{\em PushInBox}\xspace}

\newcommand{\DatasetOffReward}{\ensuremath{D_{rew}}\xspace}

\newcommand{\action}{\ensuremath{a}\xspace}
\newcommand{\state}{\ensuremath{s}\xspace}

\newcommand{\Distance}{\ensuremath{d}\xspace}

\newcommand{\DistanceMean}{\ensuremath{d^{M}}\xspace}

\newcommand{\ChangeReward}{\ensuremath{R^c}\xspace}
\newcommand{\DistanceReward}{\ensuremath{R^d}\xspace}
\newcommand{\PushBoxReward}{\ensuremath{R^p}\xspace}

\newcommand{\RewardObjectInBox}{\ensuremath{R^{ib}}\xspace}

\newcommand{\ObjSet}{\ensuremath{{\cal O}}\xspace}
\newcommand{\ActionSet}{\ensuremath{{\cal A}}\xspace}

\newcommand{\ThreshMask}{\ensuremath{{\tau_{mask}}}\xspace}
\newcommand{\ChangeMask}{\ensuremath{{c}}\xspace}
\newcommand{\Image}{\ensuremath{{\mathbf{I}}}\xspace}

\newcommand{\ObjInBox}{\ensuremath{N^{box}}\xspace}
\newcommand{\ObjOnGround}{\ensuremath{N^{gro}}\xspace}

\newcommand{\ie}{i.e.\xspace}
\newcommand{\eg}{e.g.\xspace}

%% file: introduction.tex
\section{Introduction}
\label{sec:introduction}

Pushing is an important motor skill that can serve different purposes like moving objects that are too large, 
too heavy, too distant to be grasped, or in heavily cluttered scenarios. As such, several types of pushes can help to solve many manipulation tasks. For example, pushes can be used to isolate objects, reorient them to improve the chances of grasping, bring them closer or put them into a container. Pushing can also aid perception by improving object segmentation. It is interesting to note that 25 out of the 50 manipulation tasks of the Meta-World benchmark involve
pushing \cite{yu2019metaworld}. Numerous works have investigated pushing tasks from several perspectives, e.g., creating a dataset of pushes \cite{yu2016more}
or studying and modeling their dynamics \cite{hogan2016feedback,ajay2018augmenting,suh2020surprising}. However, deciding whether to push or not, which push to perform, where, and how to do it to help manipulation remains an open problem.

\begin{figure}
\centering
\includegraphics[width=1\linewidth]{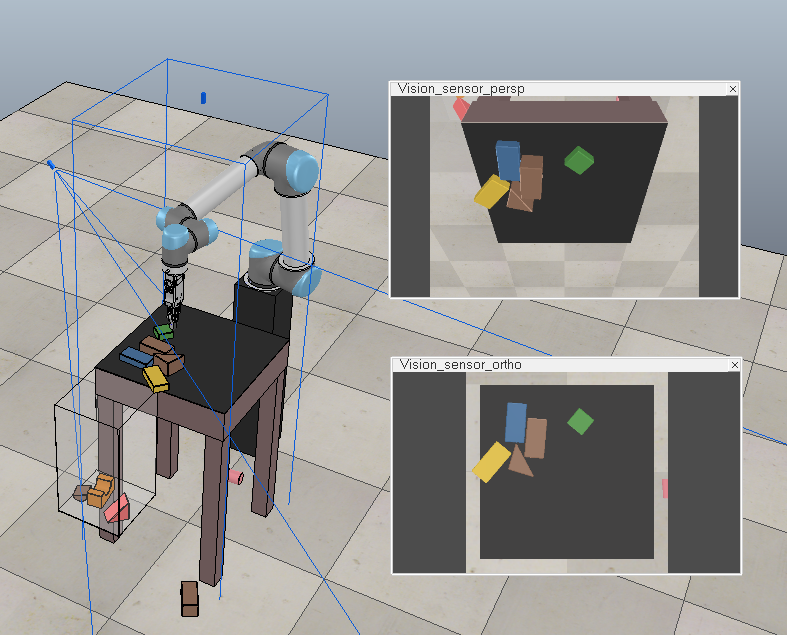}
\vspace{-2mm}
\caption{Push into Box task. The robot receives depth images as inputs and has to decide where to push so that all the objects land inside the box. The input depth images are orthographic projections of the images captured by the camera on the left. Training a Deep Neural Network to process the scene and decide which push to perform
  presents challenges, since the best pushes depend on object and box positions, as well as on the physical properties of the objects and of the table surface, which are not known by the learning agent.
}
\vspace{-6mm}
\label{fig:push_into_box_task}
\end{figure}

To address this later problem,  Zeng et al.~\cite{zeng2018learning}
proposed an interesting approach relying on a model-free deep reinforcement learning (RL)  method (DQN) to synergistically perform pushing and grasping. In their learning approach, RGB-D images are mapped using two Fully Convolutional Networks (FCNs) to a discretized action space. The first FCN maps RGB-D images to the utility of pushing 10 cm to the right at each pixel. The second FCN maps the images to the utility of grasping at each pixel. By presenting these networks with the input RGB-D image rotated at 16 different angles,
the networks can compute the utilities of pushes and grasps with different orientations. For training, Q-learning~\cite{watkins1992q} is used, relying on positive rewards for successful grasps and for pushes that produce detectable changes to the environment,
resulting in the synergistic selection of pushes and grasps. This seminal work and idea can also be found in other papers studying skill synergies,
e.g., between grasping and throwing \cite{zeng2019tossingbot},
or suction and grasping \cite{Zeng2017MultipleAffordances}. 

\begin{figure}
\centerline{
  a)\includegraphics[width=0.88\linewidth]{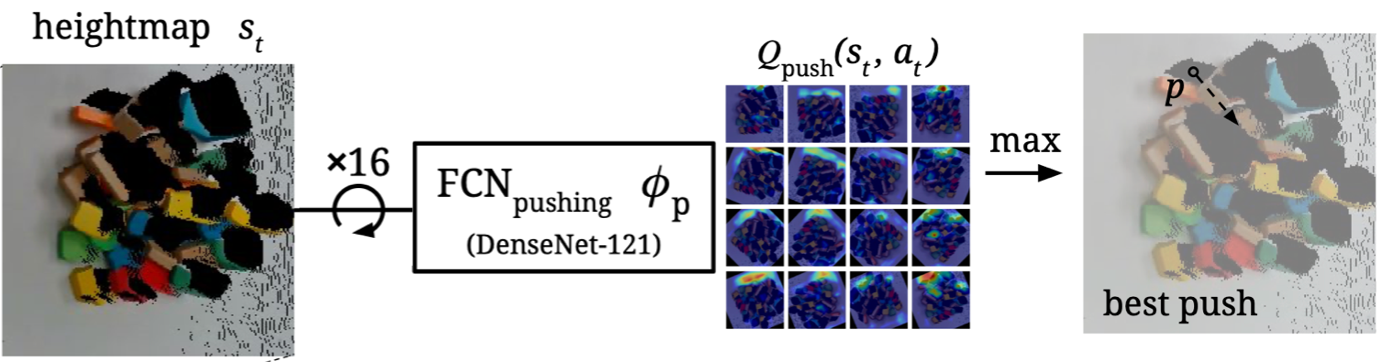}
}
\centerline{
  b) \includegraphics[width=0.26\linewidth]{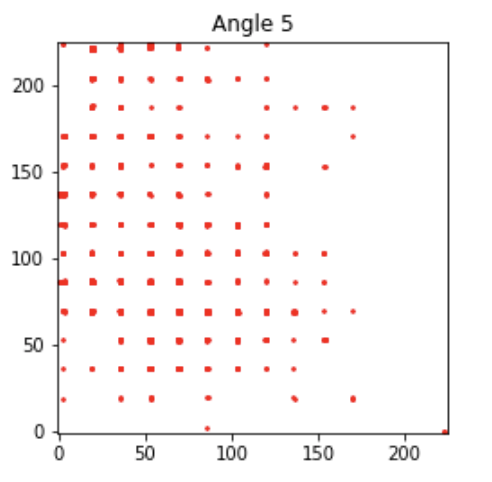}
  c) \includegraphics[width=0.18\linewidth]{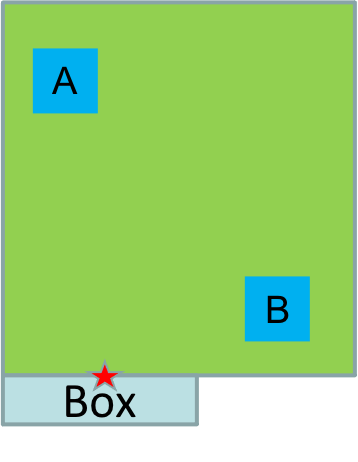}
  }
\vspace{-2mm}
\caption{a) Architecture  used in \cite{zeng2018learning} (similar in  \cite{zeng2019tossingbot,Zeng2017MultipleAffordances}),
  in which the same image classification network followed by upsampling  is applied to 16 rotations of the same input image.
  b)  Sampling locations of pushes for one direction when applying a policy learned with this architecture,
  showing the important loss in action space accuracy when applying the learned policy.
  c) Fully convolutional networks are translation (position) invariant.
  Tasks, however, may require different outputs depending on object positions.
  In the example, the cube in A should be pushed vertically,
  whereas the cube in B should be pushed horizontally.
}
\label{fig:limitations}
\vspace{-6mm}
\end{figure}

Despite their successes, these approaches present several shortcomings in their network architectures. First, they all rely on image classification network architectures
(DenseNet \cite{zeng2018learning}, ResNet \cite{zeng2019tossingbot,Zeng2017MultipleAffordances})
which progressively and drastically downscale input images before classification. In these works, bilinear upscaling is used in the last layers of their proposed networks, after the DenseNet or ResNet modules, considerably reducing the sampling accuracy of the action space when applying the policy. Even if in theory every pixel position could be sampled, selecting the action corresponding to the maximum state-action value leads to only selecting a position at the lowest resolution, as illustrated in Fig.~\ref{fig:limitations}b.

Secondly, as the {\em same} network is applied to rotated images, the network is agnostic about orientation-specific information
(it does not know which rotated image it is processing). Although this is not a problem when selecting which push better breaks objects
apart for grasping \cite{zeng2018learning}, it can be problematic when handling
task direction asymmetries,
as is the case in the task we are interested in, namely, pushing objects into a box, as illustrated in Fig.~\ref{fig:push_into_box_task}.

Thirdly, as the chosen networks are fully convolutional, they are translation invariant,
meaning that, for instance, when processing a single object on a table, they will likely predict very similar state-action values around this object whatever its position on the table. Such networks cannot handle task-dependent position asymmetries, as in our task. See Fig.~\ref{fig:limitations}c.

Finally, in \cite{zeng2018learning}, one network has to learn at the same time
which pushes are valid (\ie, alter the scene) as well as those which
are effective for the specific manipulation task (in their case, for making grasping possible). However, given the same input image and the same action parameterization, the set of valid pushes should be common to all tasks involving pushing (such as isolating objects, aligning them, bringing them closer). There is no need to learn from scratch to detect valid pushes whenever a new task involving pushing is learned. 

\mypartitle{Approach and contributions.}

In this paper, we address the task of pushing objects into a box whose environment setup is shown in Fig.~\ref{fig:push_into_box_task}. To address this task, we adopted a similar DQN learning framework as in~\cite{zeng2018learning}
to learn a pushing policy using deep convolutional neural network (CNN) agents, which entails several challenges,
as discussed above, due to orientation and position asymmetries. In this context, we first argue that this is essentially an {\em image-to-image
translation} problem where deep CNN agents predict the state-action value for all possible pushes given an input depth image. In this view, we  propose to rely on an {\em Hourglass architecture} that is much
more suitable than image classification networks  to address such problems and is also computationally more efficient than common U-Networks~\cite{ronneberger2015u}
used for image-to-image translation tasks. We also solve the orientation asymmetry challenge by
predicting state-action values for all orientations in one pass. Moreover, we propose to provide as input to the network the position of
each pixel with respect to the box in addition its depth, which could better allow the network to make decisions
based on the object location.

Secondly, in contrast to~\cite{zeng2018learning}, we introduce a {\em masking method}
that improves the sampling efficiency by constraining the set of possible
actions to those effectively leading to changes in the environment. Our driving motivation is to exploit the collaboration between two
different CNN agents, each focusing on different aspects of the task:
pushing spot detection and task-specific pushing state-action value prediction. 

Indeed, decoupling the task with two different agents brings several benefits. First, as our pushing spot detection agent has been trained to \textit{identify}
valid regions for pushing, it increases the sampling efficiency of the state-action value prediction agent
since action sampling only needs to be conducted in a much smaller set. Second, training each agent independently allows for exploiting
prior information regarding the task at hand and potentially prevents them from getting trapped in local minima as often observed with end-to-end agents. Finally, this formulation allows us to use different architectures for
the decoupled tasks, possibly increasing the complexity for the more
difficult one. In that context, our contributions are as follows:

\begin{itemize}
\item We propose to exploit the Hourglass CNN architecture in DQNs to learn object pushing policies, which leads to more efficient and more accurate high-resolution state-action value prediction and push selection.

\item We investigate the use of positional information as network input to allow position-dependent prediction behavior. 

\item We propose a decoupled approach for solving the pushing task,
  introducing  a pushing spot detection agent that works as a masking mechanism
  by identifying pushing actions leading to effective changes. Such an agent,
  once trained, can be exploited in any other manipulation
  task requiring pushes for other objectives.
\end{itemize}
We demonstrate in simulated experiments with a UR5 robot arm that our approach helps our DQN learn a pushing task faster and achieve better performance
than the state-of-the-art.


\input{related_work}

%% file: related_work.tex
\section{Related work}

For pushing, robots often rely on physics models.
These models are computable but are only approximations of physical phenomena.
For instance, Yu et al.~\cite{yu2016more} present a dataset of planar pushing experiments 
to study how reliable these models are, benchmark motion prediction methods, or for model learning.
It shows that pushing can be seen as a stochastic process even if a highly precise 
manipulator is used to perform the pushes,  highlighting the importance of learning to tackle such tasks. 
This is investigated in  \cite{ajay2018augmenting} which,  as in \cite{zeng2019tossingbot}, uses the concept of Residual Physics, i.e.,
augments analytical models with data-driven techniques to compensate for the imperfections of the models.
In \cite{ajay2018augmenting}, the trained neural networks not only correct the model predictions but also provide distributions over possible outcomes of actions.
Hogan et al.~\cite{hogan2016feedback} further address the problem of pushing an object on a plane along a desired trajectory or to pass through a sequence of via points.
The proposed method consists of using Model Predictive Control and a family of mode sequences that have been designed by the authors. The main limitation of \cite{hogan2016feedback} is the necessity of hand-designing a specific family of mode sequences for each task the robot has to solve.
In addition, all the above works only deal with single objects.

As mentioned in the introduction, Zeng et al.~\cite{zeng2018learning} use DQNs to synergistically perform pushing and grasping, 
an idea which has also been employed to explore synergies between other movements such as grasping
and throwing of arbitrary objects \cite{zeng2019tossingbot}.
In the latter, a perception module computes a feature representation of an RGB-D image, 
and a simple physics model is used to estimate the release velocity necessary to throw an object at a certain target position.
Both pieces of information (feature representation and estimated release velocity) are fed into the grasping and throwing modules.
The grasping module computes the utility of a grasp at each pixel.
The throwing module computes, for each possible grasp, the residual on top of the estimated velocity to account for phenomena
not accounted for by the simple physics model. The perception, grasping, and throwing modules are fully convolutional networks, and rewards are defined based on whether or not the thrown object landed in the correct box.
Both papers \cite{zeng2018learning,zeng2019tossingbot} suffer from limitations in their network architectures.

The methods proposed in \cite{zeng2018learning} and \cite{zeng2019tossingbot} succeed in learning to push,
grasp and throw objects in part due to the discretization of the action space, which simplifies the reinforcement learning problem.
In contrast, other methods do not require this assumption \cite{toussaint2018differentiable,strudel2019learning}.
For instance, \cite{toussaint2018differentiable} uses Logic-Geometric Programming \cite{toussaint2015logic} and Multi-Bound Tree Search \cite{toussaint2017multi}
to find and optimize a sequence of continuous actions to solve tasks involving the usage of tools and the manipulation of objects.
Notwithstanding its success, some assumptions have been made
(\eg that each object has a sphere-swept convex geometry) to make this optimization problem numerically solvable,
which may be difficult to hold in real-world applications.
We thus resorted to the DQN framework of \cite{zeng2018learning} to solve our task.

Finally, Suh and Tedrake~\cite{suh2020surprising} propose a switch-linear model to deal with the problem of pushing little carrot pieces towards a certain region. They do not deal with variable height and their images are binary, which is limiting when working with arbitrary heaps of objects. Nevertheless, they show that a switch-linear model
can surpass the performance of deep learning models for this task because the tested deep models can get stuck in a loop,
predicting actions that do not cause any change in the actual scene.
The method we propose allows us to handle this issue by introducing a module specialized in identifying pushes that lead to changes in the environment. Once trained, this module can be used in conjunction with other networks to address any pushing task.

%% file: architecture.tex
\section{Approach}
\label{sec:architecture}

\subsection{Overview}

\begin{figure*}
\centering
\includegraphics[width=0.9\linewidth]{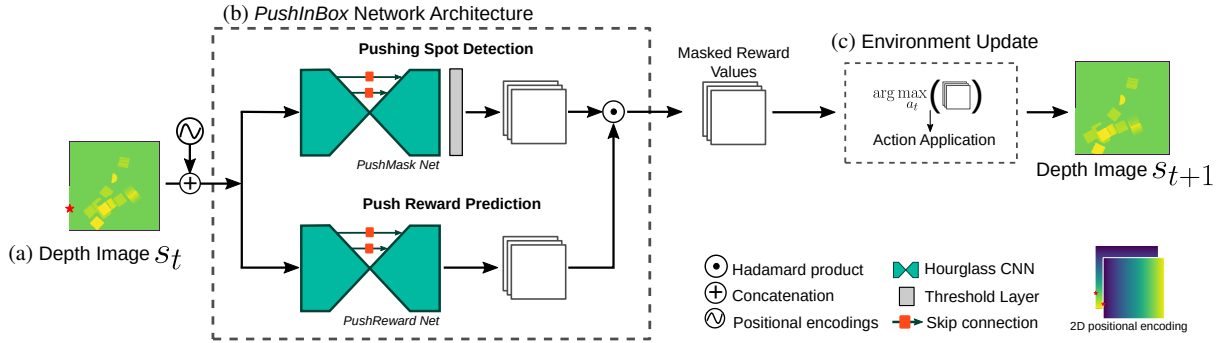}
\vspace{-2mm}
\caption{
	Proposed push into box agent collaboration pipeline.
	(a) The environment state $s_t$ (a depth image of the scene)
        is complemented with positional information and
        used as input to the \PushNet network.
        This input is processed by a \RewardNet network predicting
        the state-action value Q for each possible push,
        and by the \MaskNet predicting whether a push may lead to a scene change.
	(b) The pushing state-action values are masked by the outputs of the \MaskNet
        through elementwise multiplication to only select relevant pushes.
	(c) The environment is updated by applying the push that maximizes the
	masked state-action value maps and produces a new environment state $s_{t+1}$.
}
\label{fig:push_into_box_architecture}
\vspace{-6mm}
\end{figure*}

The proposed system is described in Fig.~\ref{fig:push_into_box_architecture}.
It takes as input a depth image,
complements this information with positional information,
and processes the result with the \emph{Push-into-Box} architecture  (termed \PushNet),
which delivers an estimate of the state-action value for each possible robot pushing action,
as parameterized by its position and orientation.
At run time, the action having the highest state-action value is executed,
and the new scene is observed. 

Looking more in detail, the \PushNet network itself is composed of two HourGlass~\cite{newell2016stacked} subnetworks:
the \MaskNet subnetwork, whose aim is to spot all the valid pushes, i.e., the pushes which result in scene changes;
and the \RewardNet subnetwork, whose aim is to predict the state-action value $Q$ that results from the reward function that we have defined to tackle the pushing task at hand.

First, we provide a more formal description of the different components of our framework, then introduce the different rewards involved in the learning, and finally the approach required for training the global network.

\subsection{Modeling}

\input{modeling}

\subsection{Rewards}
\label{sec:rewards}

\input{rewards_and_losses}

\subsection{Training protocol}
\label{sec:training}

\input{training_protocol}

%% file: modeling.tex
\mypartitle{State-space and network input.}
We represent the state $\state_t$ of the environment at time step
$t$ by its sensed depth image $\Image_t$ with shape $224\times 224$.
Additionally, the input image can be complemented with 2D positional
information of each point on the surface of the table with
respect to the box.
The image $\Image_t$ is given as input to both \MaskNet and \RewardNet subnetworks.

\mypartitle{Action-space.}
The robot can perform a pushing action  of length $\si{10}{cm}$ at any  $(x,y)$ position
in the image and in any one of $16$ directions $o$.
Hence, the action set $\ActionSet$ is the set of all the possible $(x,y,o)$ pushes
and is of dimension  $224\times 224 \times 16$. 
Each subnetwork outputs one scalar for each action. Therefore, the output of each subnetwork is composed of $16$ layers of size  $224\times 224$.

\mypartitle{Network architecture.}
We use an Hourglass architecture for each of our subnetworks to produce accurate 
and high-resolution predictions.
Each of them  comprises a U-Net-like network \cite{ronneberger2015u} that processes image features in an encoding-decoding fashion,
from high-resolution to low-resolution and back to high-resolution.
The image is first processed by a set of convolution filters
to extract a feature representation of the image.
In the downward part of the network, 
a set of residual blocks  $r_d$  progressively filters the features
from the previous resolution and downsamples the result,
until it reaches the bottleneck layers,
which correspond to an encoding of the  image at a small resolution
(128 layers at a $14 \times 14$ resolution in our case).
The upward part of the network comprises a set of residual blocks  $r_u$,
each of them processing the output from the previous (lower) resolution
followed by upsampling.
Each  $r_u$ also  integrates information from the same resolution thanks to skip connections
coming from the same resolution blocks in the downward part,
allowing to combine the different semantic levels while maintaining
the localization accuracy of the network prediction.

%% file: rewards_and_losses.tex
We introduce reward functions specific to each subtask,
i.e., valid pushing spot detection and pushing objects into the box.

\mypartitle{Pushing spot detection reward.}
We design this reward under the premise that pushing actions
should produce \emph{effective changes}. 
More precisely, given the environment  state at time $t$, $\state_t$ (the depth image $\Image_t$), 
and the executed action $\action_t$ leading to the next state $\state_{t+1}$ (image  $\Image_{t+1}$),
we define a notion of change by  counting the amount of local pixel changes
\begin{equation}
  \label{eq:change}
  \ChangeMask_t = \sum_{p}  I_{(|\Image(p)_t - \Image(p)_{t+1}| >= \si{1}{cm})}, 
\end{equation}
where $I_v$ is the indicator function, equal to 1 if $v$ is true, and 0 otherwise.
Accordingly, we define the \MaskNet change reward
\begin{equation}
  \label{eq:changereward}
\ChangeReward \left(\action_t, \state_t, \state_{t+1}\right) = \begin{cases}
		0 & \text{ if }  \ChangeMask_t < \ThreshMask
		\\
		1 &  \text{otherwise} 
	\end{cases},
\end{equation}
where $\ThreshMask$ is a threshold which we set to 1000 in practice.
%

\mypartitle{Push-into-box reward.}
We define an effective push as that which moves the objects towards the box,
and hence will minimize the average distance between object pixels and the box. 
Accordingly,  our measure of pushing effectiveness is defined as:
\begin{equation}
  \label{eq:distance}
  \Delta\DistanceMean_{t} = \DistanceMean_{t} - \DistanceMean_{t+1} \mbox{ with }   \DistanceMean_{t} = \frac{1}{|\ObjSet_t|} \sum_{p \in \ObjSet_t} \Distance_{t}(p), 
\end{equation}
where $p$ is a pixel position in the image, 
$\Distance_{t}(p) = \| p - p_{box} \|$ denotes the distance between
pixel $p$ and the center of the edge of the box in contact with the table $p_{box}$, 
 $\ObjSet_t$ is the set of objects' pixels obtained by 
thresholding the scene depth image,  and $|\ObjSet_t|$ denotes its cardinality.
Finally we define the push-into box reward \PushBoxReward as 
the sum of a distance reward
$\DistanceReward_t = \max \left(0,  \Delta\DistanceMean_{t} \right)$
measuring how much objects move closer to the box, as well as a
reward $\RewardObjectInBox_t = 10 \times\ObjInBox_t$ proportional
to the number of objects  $\ObjInBox_t$ falling into the box,
unless one or more objects fall to the ground ($\ObjOnGround_t > 0$)
where $\ObjInBox_t$ and $\ObjOnGround_t$  are provided by the environment simulator.
More formally,
\begin{equation}
\PushBoxReward\left(\action_t, \state_t, \state_{t+1}\right) = \begin{cases}
		0 & \text{if } \ObjOnGround_t > 0\\
		\DistanceReward_t + \RewardObjectInBox_t & \text{otherwise}
	\end{cases}.
\end{equation}

%% file: training_protocol.tex
In this section, we indicate the protocol followed to train our architecture,
which involves both supervised training and reinforcement learning steps.
The main steps are as follows:
\begin{enumerate}
\item Train \MaskNet using appropriate data. The parameters of this network are assumed to be frozen afterward.
\item Train \RewardNet using an offline dataset \DatasetOffReward.
\item Train the whole \PushNet using online reinforcement learning, taking advantage of the \MaskNet to improve training. In this step, the parameters of \MaskNet are frozen.
\end{enumerate}
Before describing the specific aspects of how the \MaskNet, \RewardNet, and \PushNet models
were trained, we introduce the basic elements of reinforcement learning.

\mypartitle{Reinforcement learning (RL).}
During RL, the system simulates a set of $B$ experiences, which can be expressed as triplets $(\state_t,\action_t,\state_{t+1})$.
For each experience, we can compute the expected state-action value $Q(a_t, s_t)$ as predicted by the network.
The output $Q(a_t, s_t)$ should be as close as possible to the target value
\begin{align}
	&y\left(a_t, s_t, s_{t+1}\right) = \nonumber\\
	&R\left(a_t, s_t, s_{t+1} \right) + \gamma Q \left( s_{t+1}, \underset{a'}{\arg\max}\left(Q\left(a', s_{t+1}\right)\right) \right),
	\label{eq:target_value}
\end{align}
which can be computed from the definition of the reward and $\underset{a'}{\arg\max}\left(Q\left(a', s_{t+1}\right)\right)$. The latter can be computed by applying the network to the state $\state_{t+1}$ of the current experience.
In other words, for an experience $i$, the temporal difference error $\delta^i_t = L_e(Q\left(a^i_t, s^i_t\right),  y\left(a^i_t, s^i_t, s^i_{t+1}\right))$ must be minimized,
where $L_e$ is the loss used to measure the discrepancy between the prediction and the target. 
The loss for each mini-batch of size $B$ is $Loss = \sum_{i=1}^B \delta^i_t$, which can be optimized via gradient descent optimization algorithms and using backpropagation to compute the gradients.
We relied on the Adam optimizer. 

\begin{figure}
\centering
\includegraphics[width=1.0\linewidth]{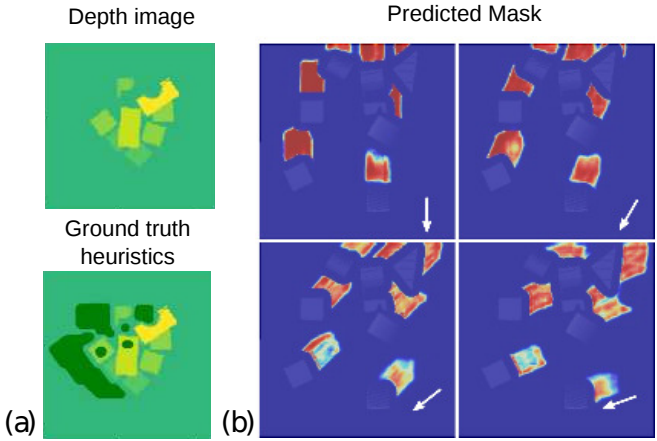}
\caption{
	(a) Example of generating ground truth with our proposed heuristics.
		Top: Input depth image; Bottom: generated ground truth push mask for pushes pointing to the right.
	(b) Example of generated valid pushing probability maps predicted by our \MaskNet for different pushing orientation possibilities on a sample image. The color red indicates start positions of pushes with a higher probability of leading to changes, while blue indicates a lower probability of leading to changes.
}
\label{fig:mask_heuristics}
\vspace{-6mm}
\end{figure}

\mypartitle{Training \MaskNet.}
To train the \MaskNet we first generate a training set $D_{m}$ using a heuristics
that identifies pixels close to the borders of objects as promising or not to push according to edge and orientation analysis. 
More precisely, we first generate a set of scenarios by tossing 10 randomly selected objects 
onto the table and generating their corresponding depth images.
To compute the ground truth, we start by detecting edges using the Canny edge detector.
Then, we label an edge pixel $p$ to be a valid pushing position if $\Image(p_{\theta}) - \Image(p) > \delta$, where $p_{\theta}$ is the pixel $p$ moved in a line along the pushing 
orientation possibility $\theta$. The reasoning behind this procedure is that a push starting at $p$ and passing through $p_{\theta}$ will probably meet an object border if $p_{\theta}$ is higher than $p$ by a certain threshold $\delta$, meaning that this push will probably produce some significant change.
Fig.~\ref{fig:mask_heuristics}(a) shows an example of the ground truth generated 
with this heuristics for a given image.

Our goal is to train a network, \MaskNet, which identifies whether a given push $\action$ is leading to a change or not.
In this regard, we have a binary classification problem for each $\action$,
and we thus rely on the binary cross-entropy loss  $L_e$ for training.
The network is trained in two stages.
In the first stage, we train \MaskNet using the dataset $D_{m}$,
i.e., the ground truth masks generated by the described heuristics.
However, as this heuristics does not account well for setup characteristics (gripper shape and size,
friction, noise), we run a refinement training stage (in an RL fashion) that uses actual object pushes
and Eq.(\ref{eq:changereward}) as ground truth for supervised learning.
Fig.~\ref{fig:mask_heuristics}(b) illustrates our results on a sample image.

\mypartitle{Training \RewardNet and \PushNet.}

We divide the \PushNet training into two stages.
In each stage, the network is trained to minimize the temporal difference errors $\delta_t^i$.

In the first stage (offline RL)
we train \RewardNet from scratch from a dataset $\DatasetOffReward$ of  experiences obtained as follows:
scenarios are generated at random, and the pushing actions are sampled uniformly at random 50\% of the time,
or using a trained \MaskNet otherwise.
Specifically, in the latter case, we compute the forward pass of the \MaskNet on the current
depth image and generate a binary mask of pushing actions.
Then, we select pushing actions that lead to a change in the environment with
probability greater than a certain threshold.
These actions are executed in simulation and the corresponding experiences are recorded. We have chosen the probability threshold to select the actions based on the F-score obtained by different thresholds in a test set, as detailed in Section \ref{sec:experiments}.
Subsequently, \RewardNet is trained to minimize the temporal difference errors associated with the recorded experiences.

In the second stage (online RL) we further train \RewardNet in simulation,
using the previously trained \RewardNet model as the starting point and exploring new experiences with an epsilon-greedy approach.
We proceed as follows:
given each new depth image experienced in simulation, we compute the forward pass of \RewardNet 
and \MaskNet.
These predictions are then combined via the Hadamard product.
After each new batch of experiences, the parameters of \RewardNet are optimized to minimize the squared error loss $L_e$ associated with the experiences in this new batch.

The offline RL stage allows for an effective initialization of \RewardNet with pre-recorded experiences while the online RL stage further improves the network by exploring new experiences. In this work, for simplicity, \MaskNet is trained and its parameters are frozen before training \RewardNet. One could also consider training \MaskNet further together with \RewardNet.

%% file: experiments.tex
\section{Experiments}
\label{sec:experiments}

\subsection{Experimental protocol}
\label{sec:experimental_protocol_mask}

\mypartitle{Experimental protocol for \MaskNet.}
As this network is trained independently, we generated a test dataset to evaluate the performance of the different models.
To that end, we generated a test set by randomly sampling 10240 scenarios involving 1 to 10 objects, along with random pushes.
The ground-truth (GT) regarding these pushes has been determined as during training
(i.e., thresholding the number of pixels that have changed after the push).
This test set was used to produce precision-recall curves.
The precision is defined as: out of the set of pushes identified by the system as valid (true),
what is the proportion of pushes that are true (according to the GT). The recall corresponds to the proportion of true pushes correctly
identified by the system. 

\mypartitle{Push-into-box experimental protocol.}
We created 100 random scenarios comprising exactly 10 objects.
For each model under evaluation, we applied the corresponding policy,
until either there was no more object on the table, or the policy did not produce any change 10 times in a row (implicitly implying that the model is pushing
in the void).

\mypartitle{Evaluation Metrics.}
We quantify the performance of our models by the mean and standard deviation
of the number of objects successfully pushed into the box \NObjBox, number of 
objects that fell on the ground \NObjGround, and number of objects left on 
the table \NObjTable.
Additionally, we measure the efficiency of our system by computing the ratio between the average number of objects successfully pushed into the box \NObjBox and the average number of executed actions \Naction.

\mypartitle{Tested models.}
To evaluate the benefits of the different parts of our framework, we tested different algorithmic configurations. These elements are specified when reporting results in Table~\ref{Tab:test_10_objects}.
\begin{itemize}
\item \Baseline architecture \cite{zeng2018learning}: in this case, we used the architecture proposed by \cite{zeng2018learning} as \RewardNet network,
  which is based on a DenseNet image recognition network pre-trained on ImageNet \cite{deng2009imagenet}, with an improvement.
  Rather than predicting the reward for the 16 pushing directions by rotating the image 16 times and applying the
  DensetNet architecture, we modified the network so that it can predict the reward for all directions in the last stage of the network,
  as is done with our architecture. 
\item \HourGlass architecture: for the \RewardNet network. 
\item \HourGlassLPE architecture: the same as \HourGlass, but using an additional position encoding for the network input. 
  
\item \WithMask option: this indicates whether the \MaskNet network was exploited by the system,
  both during training in the online RL step and in testing.
\item \WithRL: whether the \RewardNet network was only trained with the offline dataset $\DatasetOffReward$ 
  or if an additional RL fine-tuning with an epsilon-greedy exploration strategy was conducted (\WithRL = Yes). 
\end{itemize}

\
\subsection{Results}
\label{sec:results}

\begin{figure}
\centering
\includegraphics[width=0.8\linewidth]{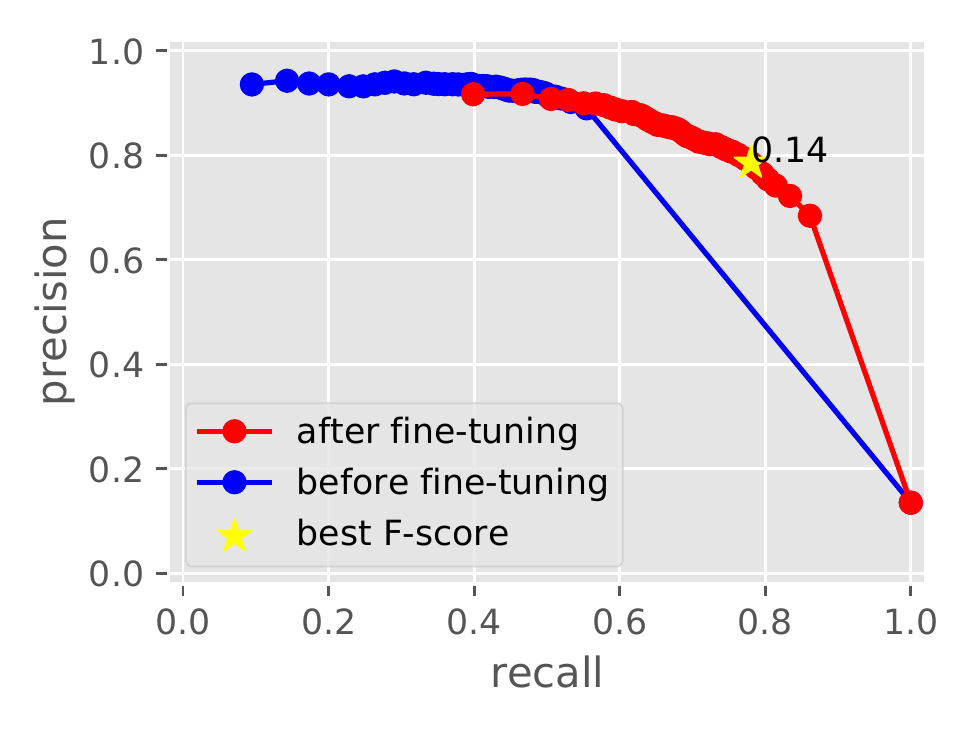}
\vspace{-4mm}
\caption{Precision vs. recall curves of \MaskNet before fine-tuning (trained only based on a heuristics) vs. after fine-tuning with actual pushes. The model with probability threshold $0.14$ obtained the best F-score.
}
\label{fig:precision_vs_recall}
\vspace{-2mm}
\end{figure}

\begin{figure}
\centerline{
  \includegraphics[width=1\linewidth]{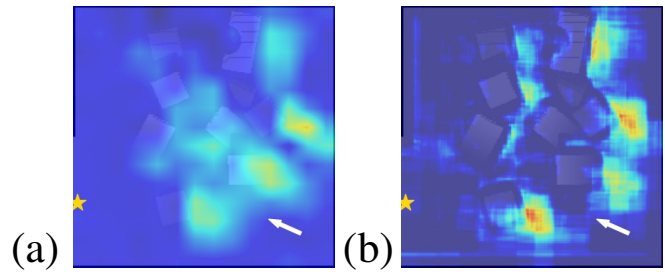}
  }
\vspace{-2mm}
\caption{
	Comparison of output state-action value maps for (a) Densenet, and
	(b) Hourglass architectures. 
	The smoothing effects in (a) appear by producing the maps at very low
	resolution.
	A more precise localization of pushing action can be identified in the maps produced
	by the Hourglass network in higher resolution.
}
\label{fig:denset_vs_hg}
\vspace{-6mm}
\end{figure}

\mypartitle{\MaskNet results.}
As shown in Fig.~\ref{fig:precision_vs_recall}, training based only on the masks generated by the heuristics led
to high precision but resulted in missing some of the pushes which produced changes (low recall).
In this regard, the \MaskNet benefited from the fine-tuning with actual pushes.
The fine-tuning led to a higher recall and a higher F-Score.
The probability threshold of 0.14 to detect changes has achieved the highest F-Score
and has been used in our further experiments involving the \MaskNet.

\begin{figure}
\centerline{
 a) \includegraphics[width=1\linewidth]{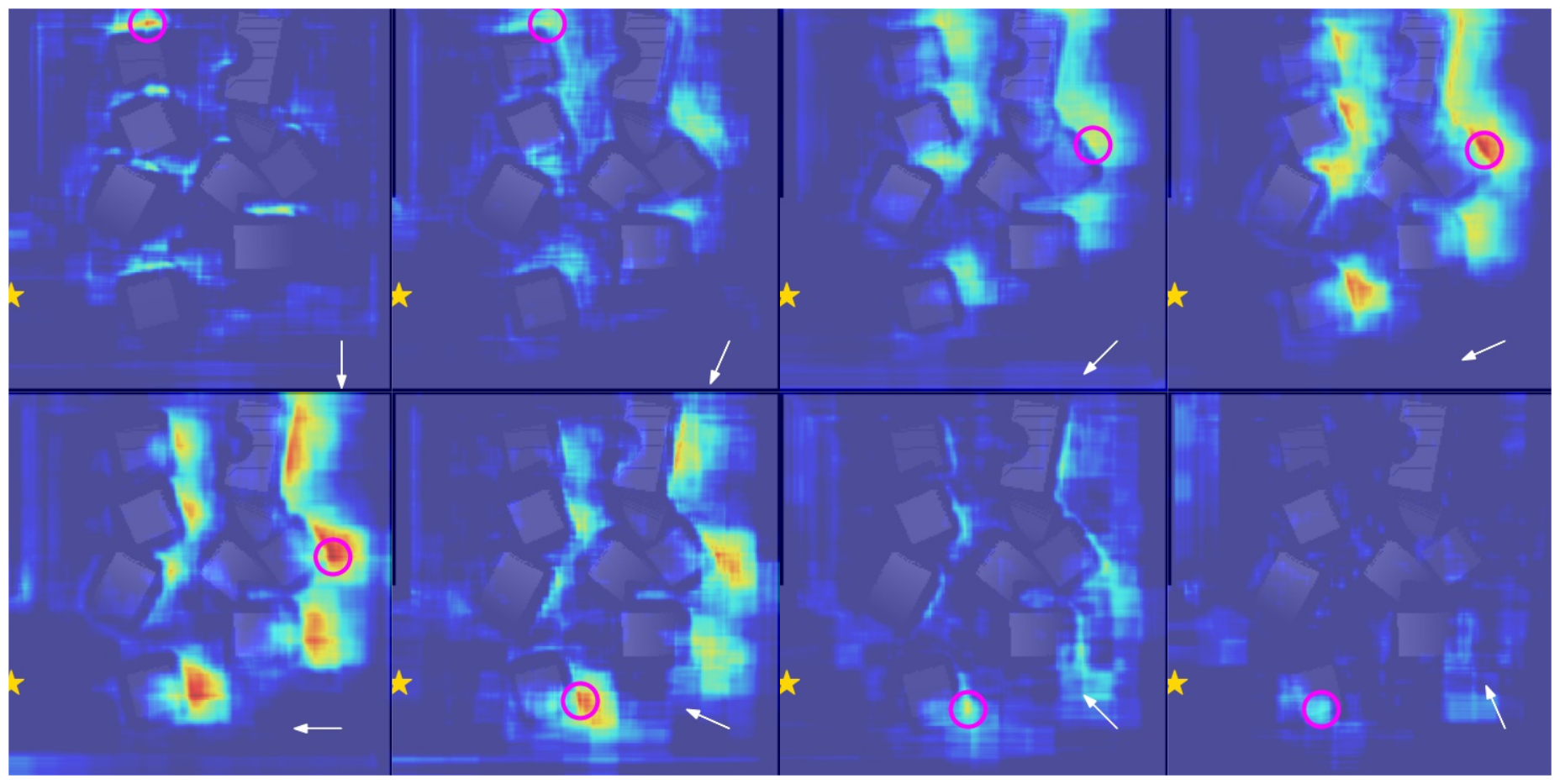}
  }
\vspace{2mm}
\centerline{
  b) \includegraphics[width=1\linewidth]{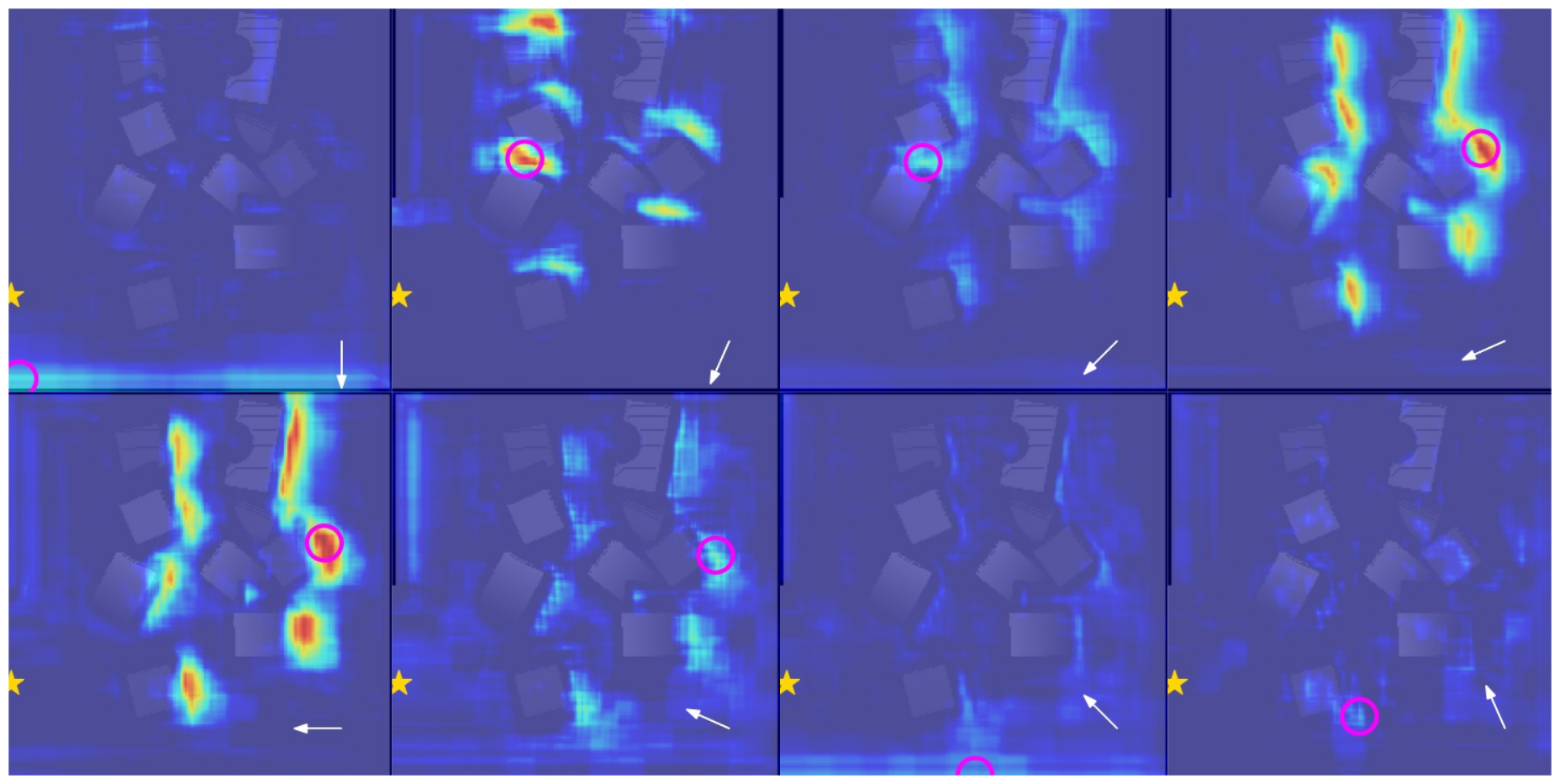}
  }
\vspace{-2mm}
\caption{
  Examples of state-action value map predictions for a set of pushing directions.
  a)  \HourGlass architecture, using the \MaskNet network, and \WithRL with $\gamma=0$.
  b) Same as a), but with $\gamma=0.4$.
  A white arrow shows the direction of the pushing action, the magenta circle
  center indicates the location of the maximum for that state-action value map and the golden star indicates the position of the center of the edge of the box in contact with the table.
}
\label{fig:reward_pred}
\vspace{-6mm}
\end{figure}

\mypartitle{\PushNet results.}
Our main quantitative results are summarized in Table~\ref{Tab:test_10_objects}. The best result in each column is bold, the second-best, underlined.
All the HourGlass-based models have surpassed the DenseNet-based ones in terms of the average number of objects correctly pushed into the box. Figs.~\ref{fig:denset_vs_hg} and \ref{fig:reward_pred} show state-action value predictions made by different models.

Comparing the \HourGlass models with and without the mask after online RL, we can observe that the mask only slightly
improved the performance in terms of the average number of objects in the box and efficiency (fewer actions).
Nevertheless, the mask has given the  \HourGlassLPE model a significant boost in performance before online RL
($8.05$ objects on average in the box with the mask versus $6.71$ without).
We hypothesize that the mask helps to prevent models from
sliding on the top of objects or pushing where there is no object.

The model with the highest performance overall has been the \HourGlass with mask and after online RL with discount factor $\gamma=0$.
By observing the behavior of this model during tests, we noticed that the main sources of problems preventing the model from pushing all the objects into the box are objects very close to the borders of the table, where the robot is unable to move them back
to the center due to its limited workspace, and cylinders, which easily roll out of the table.

We had expected models with discount factor $\gamma=0.4$ to perform better than the models with $\gamma=0$,
but this has not been the case.
A possible explanation is that this task does not require learning complex successions of pushes to obtain higher expected future rewards.

Finally, while the model with linear position encoding performed best in offline learning,
exhibiting a better behavior for carefully bringing objects in front of the box before pushing them into the box,
its performance after online RL decreased.
A closer look at the learning curve showed a relatively chaotic behavior beyond
a certain number of epochs, suggesting that the
selected optimizer and parameters are not optimal.
This will be investigated in future work.

\begin{table*}[t]
	\caption{
    	Push into the box performance results obtained by testing our model in scenarios with 10 objects
    }
	\centering
    \begin{tabular}{lccllllc}
        \toprule
      Architecture &   \WithRL & \WithMask  & \NObjBox (std) & \NObjGround (std) & \NObjTable (std) & \Naction & $\frac{\mbox{avg.} \NObjBox}{\text{avg.} \Naction}$ \\
        \addlinespace[1pt]
      \midrule
      \Baseline & No & No  & $4.56 (2.1) $ & $4.38 (1.7)$ & $1.06 (2.2)$ & $\mathbf{18.0} (8.4)$ & $0.25$ \\
      \Baseline & Yes & No  & $5.38 (3.2) $ & $1.17 (1.2)$ & $3.45 (3.9)$ & $21.5 (8.4)$ & $0.25$ \\ \midrule 
      \HourGlass & No & Yes   & $7.80 (1.2) $ & $2.15 (1.1) $ & $0.05 (0.4)$ & $23.7 (4.8) $ & $0.33$ \\ 
      \HourGlass & No & No   & $7.76 (1.4) $ & $1.90 (1.3) $ & $0.34 (1.2)$ & $26.6 (9.6) $ & $0.29$ \\ 
      \HourGlassLPE & No  & Yes   & $8.05 (1.7) $ & $1.48 (1.2) $ & $0.47 (1.6)$ & $28.2 (10.8) $ & $0.29$ \\
      \HourGlassLPE & No & No &  $6.71 (2.6) $ & $1.21 (1.3) $ & $2.08 (3.0)$ & $30.0 (12.6) $ & $0.22$ \\  \midrule
      \HourGlass, $\gamma=0$ & Yes & Yes &  $\mathbf{8.96} (0.9) $ & $\underline{1.01} (0.9) $ & $\underline{0.03} (0.2)$ & $21.5 (4.6) $ & $\mathbf{0.42}$ \\
      \HourGlass, $\gamma=0$ & Yes  & No &  $\underline{8.94} (1.0) $ & $\mathbf{1.00} (1.0) $ & $0.06 (0.4)$ & $23.2 (6.7) $ & $\underline{0.39}$ \\
      \HourGlass, $\gamma=0.4$ & Yes & Yes &  $8.00 (1.4) $ & $1.98 (1.4) $ & $\mathbf{0.02} (0.1)$ & $\underline{20.6} (6.9) $ & $\underline{0.39}$ \\
      \HourGlass, $\gamma=0.4$ & Yes & No &  $7.61 (1.6) $ & $2.11 (1.2) $ & $0.28 (1.4)$ & $21.1 (5.3) $ & $0.36$ \\
      \HourGlassLPE, $\gamma=0$ & Yes & Yes   & $6.66 (1.8) $ & $2.89 (1.4) $ & $0.45 (1.8)$ & $24.2 (7.0) $ & $0.28$ \\ 
      \HourGlassLPE, $\gamma=0.4$ & Yes & Yes &  $6.59 (1.7) $ & $3.16 (1.5) $ & $0.25 (1.3)$ & $24.0 (8.8) $ & $0.27$ \\ \midrule 
      \bottomrule
    \end{tabular}
    \vspace{1em}
    \label{Tab:test_10_objects}
    \vspace{-6mm}
\end{table*}

